\documentclass{article}

\PassOptionsToPackage{numbers, compress}{natbib}

\usepackage[preprint]{neurips_2021}

\usepackage[utf8]{inputenc} 
\usepackage[T1]{fontenc}    
\usepackage{url}            
\usepackage{booktabs}       
\usepackage{amsfonts}       
\usepackage{nicefrac}       
\usepackage{microtype}      
\usepackage{xcolor}         
\usepackage{ulem}
\usepackage{amsmath}
\usepackage{amssymb}
\usepackage{bbm}
\usepackage{graphicx}
\usepackage{subfigure}
\usepackage{multirow}
\usepackage{arydshln}

\usepackage{algorithm}
\usepackage{algorithmic}

\title{Improving White-box Robustness of Pre-processing Defenses via Joint Adversarial Training}

\author{
    Dawei Zhou \\
    Xidian University\\
    \And
    Nannan Wang \\
    Xidian University\\
    \And
    Xinbo Gao \\
    Chongqing University of\\Posts and Telecommunications\\
    \And
    Bo Han \\
    Hong Kong \\ Baptist University \\
    \And
    Jun Yu \\
    University of Science and\\Technology of China\\
    \And
    Xiaoyu Wang \\
    The Chinese University of\\Hong Kong (Shenzhen)\\
    \And 
    Tongliang Liu \\
    The University of Sydney\\
}
\begin{document}

\maketitle

\begin{abstract}
Deep neural networks (DNNs) are vulnerable to adversarial noise. A range of adversarial defense techniques have been proposed to mitigate the interference of adversarial noise, among which the input pre-processing methods are scalable and show great potential to safeguard DNNs. However, pre-processing methods may suffer from the \textit{robustness degradation effect}, in which the defense reduces rather than improving the adversarial robustness of a target model in a white-box setting. A potential cause of this negative effect is that adversarial training examples are static and independent to the pre-processing model. To solve this problem, we investigate the influence of \textit{full adversarial examples} which are crafted against the full model, and find they indeed have a positive impact on the robustness of defenses. Furthermore, we find that simply changing the adversarial training examples in pre-processing methods does not completely alleviate the robustness degradation effect. This is due to the adversarial risk of the pre-processed model being neglected, which is another cause of the robustness degradation effect. Motivated by above analyses, we propose a method called \textit{Joint Adversarial Training based Pre-processing} (JATP) defense. Specifically, we formulate a feature similarity based adversarial risk for the pre-processing model by using full adversarial examples found in a feature space. Unlike standard adversarial training, we only update the pre-processing model, which prompts us to introduce a pixel-wise loss to improve its cross-model transferability. We then conduct a joint adversarial training on the pre-processing model to minimize this overall risk. Empirical results show that our method could effectively mitigate the robustness degradation effect across different target models in comparison to previous state-of-the-art approaches.

\end{abstract}

\section{Introduction}
Although deep neural networks (DNNs) achieve great success in applications such as computer vision \cite{li2014common,he2016deep,li2016mutual,2017Mask,dosovitskiy2020image,bai2022improving}, speech recognition \citep{wang2017residual} and natural language processing \citep{sutskever2014sequence,devlin2018bert}, they are found to be vulnerable to adversarial examples which are crafted by adding imperceptible but adversarial noise on natural examples \citep{szegedy2013intriguing,goodfellow2014explaining,tramer2020adaptive,wang2022triangle}. Adversarial examples can remain destructive in the physical world \citep{wu2019defending,duan2020adversarial} and transfer across different models \citep{wu2020skip,huang2019black}. The vulnerability of DNNs raises security concerns about their reliability in decision-critical deep learning applications, e.g., autonomous driving \citep{eykholt2018robust} and person recognition \citep{tang2004video,liu2006spatio,gong2013multi,xu2020adversarial}.

A major class of adversarial defenses pre-process adversarial examples to mitigate the interference of adversarial noise without modifying the target model \citep{xu2017feature,feinman2017detecting,liao2018defense}. Among pre-processing methods, compared with feature squeezing \citep{guo2017countering,xu2017feature} and adversarial detection \citep{ma2018characterizing,qin2019detecting}, input denoising \citep{liao2018defense,shen2017ape,samangouei2018defense,schott2018towards,ghosh2019resisting,lin2019invert,naseer2020self} can remove adversarial noise and sufficiently retain the original information of the natural examples. Thus, input denoising methods are widely researched and they have shown great potential to safeguard DNNs from adversarial attacks.

\begin{figure}[t]
    \centering
    \subfigure[Standard]{
        \label{fig1_1}
        \includegraphics[width=1.7in]{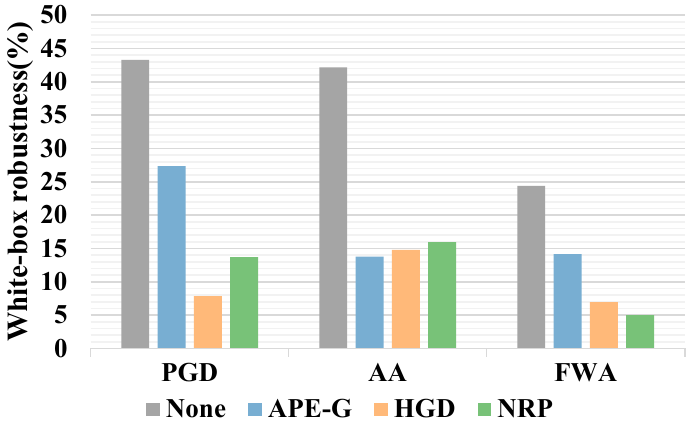}}
    \hspace{0in}
    \subfigure[TRADES]{
        \label{fig1_2}
        \includegraphics[width=1.7in]{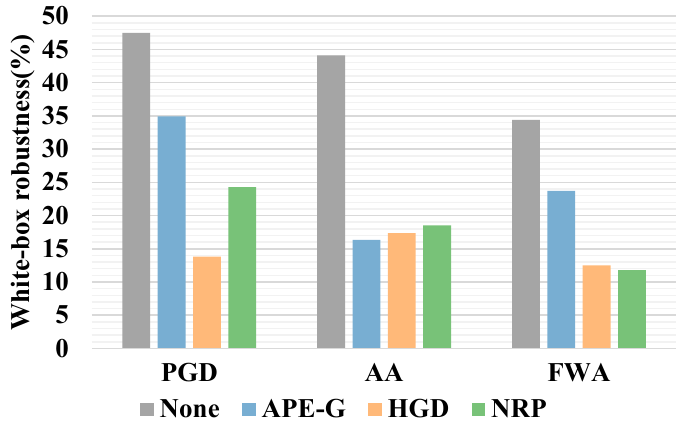}}
    \hspace{0in}
    \subfigure[Comparison on TRADES]{
        \label{fig1_3}
        \includegraphics[width=1.7in]{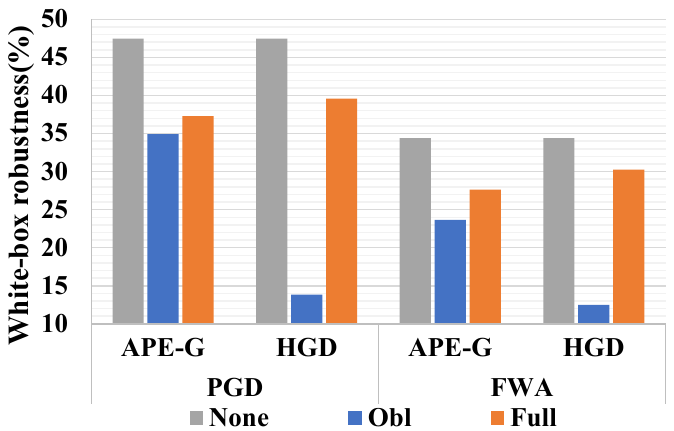}}
    \caption{The visualization of robustness degradation effect. We evaluate the white-box robustness (accuracy on white-box adaptive attacks) of three pre-processing defenses: APE-G \citep{shen2017ape}, HGD \citep{liao2018defense} and NRP \citep{naseer2020self} on \textit{CIFAR-10} \citep{krizhevsky2009learning}. The target models are adversarially trained via two adversarial training strategies: \textit{Standard} \citep{madry2017towards}, and \textit{TRADES} \citep{zhang2019theoretically}. We combine the adaptive attack strategy with three attacks such as PGD \citep{madry2017towards}, AA \citep{croce2020reliable} and FWA \citep{wu2020stronger} to craft adversarial examples. "None" denotes that no pre-processing defense is used. "Obl" denotes the pre-processing model trained using oblivious adversarial examples, and "Full" denotes the model trained using full adversarial examples.}
    \label{fig1}
\vskip -0.1in
\end{figure}

However, many denoising methods only present security against \textit{oblivious attacks} \citep{athalye2018robustness,tramer2020adaptive} where an attacker is completely unaware the defenses is being applied. Unfortunately, security against oblivious attacks is far from sufficient to be useful in practice. A serious attacker would certainly consider the possibility that a defense is used \citep{athalye2018robustness}. Recent researches \citep{carlini2017magnet,tramer2020adaptive} show that effectiveness of denoising methods are significantly weakened in an adaptive threat model where an attacker has full access to the pre-processing model or can estimate the knowledge (e.g., gradient) of the pre-processing model \citep{carlini2017towards}. Especially, for a robust target model which is adversarially trained via state-of-the-art adversarial training strategies \citep{madry2017towards,ding2019sensitivity,wang2019convergence,zhang2019theoretically}, empirical results as shown in Fig~\ref{fig1_1},~\ref{fig1_2} present that using the denoising defenses to pre-process inputs may significantly reduce rather than improve the adversarial robustness against worst-case (\textit{i.e.,} white-box) adaptive attacks. We name such adversarial robustness as \textit{white-box robustness}, and this phenomenon as “\textit{robustness degradation effect}”.

A potential cause of the negative effect is that adversarial training examples are static and independent to the pre-processing model. Recall that adversarial examples used in pre-processing methods are typically crafted only against the target model. Since the target model is pre-trained and their parameters are fixed, these \textit{oblivious adversarial examples} specific to the target model are also fixed and not associated with the pre-processing model. From this perspective, using such training examples cannot guarantee that the defense model can effectively deal with the adaptive attacks. To solve this problem, a natural idea is to use \textit{full adversarial examples} which are crafted against the full model (a single model composed of a pre-processing model and a target model) to replace oblivious adversarial examples. We investigate the influence of full adversarial examples on the robustness and find that they exhibit a better impact compared to oblivious adversarial examples. Specifically, we conduct a proof-concept experiment in a white-box setting. As shown in Fig~\ref{fig1_3}, compared to the original pre-processing models, the models trained by full adversarial examples achieve higher accuracy. The details of this experiment are presented in Section~\ref{section2.2}. This discovery inspired us to use full adversarial examples to train the pre-processing model. In addition, to prevent the \textit{label leakage} \citep{kurakin2016adversarial,zhang2019defense} from affecting the defense generalization to unseen attacks, we explore to make full adversarial examples independent to the label space by maximally disrupting deep features of natural examples on an internal layer of the full model.


Note that the results in Figure~\ref{fig1_3} also show that simply modifying the adversarial training examples in pre-processing methods does not completely alleviate the robustness degradation effect. This may be caused by the vulnerability of the pre-processing model. The denoising defenses typically exploit a generative network to train the pr-processing model for recovering adversarial examples. Unfortunately, existing works \citep{gondim2018adversarial,kos2018adversarial,chen2020breaking,sun2020type} have demonstrated that classic generative models are vulnerable to adversarial attacks. An attacker can mislead the pre-processing model to generate an output with respect to wrong class by disrupting the recovered example as much as possible. The pre-processing methods typically only focus on the ability of the pre-processing model to remove oblivious adversarial noise, but overlook the risk of the model being perturbed, which results in their insufficient adversarial robustness. To address this issue, we formulate an adversarial risk for the pre-processing model, to exploit the full adversarial examples to improve the inherent robustness of the pre-processing model instead of just using them to learn the denoising mapping. Corresponding to the above full adversarial examples found in feature space, we introduce a feature similarity term to measure the distance between natural and full adversarial examples. By this design, the adversarial risk is expected to reduce the distortion in both label space and feature space. The details of the adversarial risk are presented in Section~\ref{section2.3}.


Motivated by the above analyses, we propose a \textit{Joint Adversarial Training based Pre-processing} (JATP) defense. Specifically, we use full adversarial examples found in the feature space as the supervision signal to train the pre-processing model. Then, we formulate a feature similarity based adversarial risk for the pre-processing model to improve its inherent robustness by using above full adversarial examples. Note that unlike the standard adversarial training which is model independent, our method only updates the parameters of the pre-processing model and needs to ensure that the parameters are suitable for different target models. This requires us to improve the cross-model transferability. Considering that the natural examples used by different target models are consistent, and they have no adversarial patterns, we therefore introduce a pixel loss to reduce the distance between the pre-processed and natural examples. Based on the above designs, we conduct a joint adversarial training on the pre-processing model to minimize this overall risk in a dynamic manner. Experimental results in Section~\ref{section4} show that our method effectively mitigates the robustness degradation effect against unseen types of white-box adaptive attacks in comparison to previous pre-processing defenses. 

The main contributions in this paper are as follows:

\begin{itemize}
    \item We analyze two potential factors that cause the robustness degradation effect: (1) adversarial training examples used in pre-processing methods are independent to the pre-processing model; and (2) the inherent robustness of the pre-processing model is not sufficient due to its adversarial risk is neglected during the training process.
    \item we first formulate a feature similarity based adversarial risk for the pre-processing model to improve its inherent robustness by using adversarial examples crafted in a feature space against the full model. Then, we introduce a pixel-wise loss to improve the cross-model transferability of the pre-processing model. We propose a \textit{Joint Adversarial Training based Pre-processing} (JATP) defense to minimize the overall risk in a dynamic manner.
    \item Experimentally, we demonstrate that JATP defense could significantly improve the white-box robustness of pre-processing defenses against adaptive attacks and mitigate the robustness degradation effect compared with the state-of-the-art. In addition, it could be applied to safeguard across different target models without additional training procedures.
\end{itemize}

The rest of this paper is organized as follows. In Section~\ref{section2},
we analyze the robustness degradation effect. In Section~\ref{section3}, we describe our defense method and present its implementation. Experimental results on different datasets are provided in Section~\ref{section4}. Finally, we conclude this paper in Section~\ref{section5}. In addition, we briefly review related work on attacks and defenses in supplementary material A.

\section{Analyzing the robustness degradation effect}
\label{section2}
In this section, we analyze two potential causes of the robustness degradation effect and explore to find their solutions. 

\subsection{Preliminaries}
\label{section2.1}
We first define the notation. We use \textit{bold lower-case} letters (e.g., $\mathbf{x}$, $\mathbf{y}$) and \textit{lower-case} letters (e.g., $x$, $y$) to denote vectors and scalars respectively. We use \textit{upper-case} calligraphic symbols such as $\mathcal{T}$ and  $\mathcal{P}$ to denote models.

In the setting of a $K$-class ($K\ge2$) classification problem, we are given a dataset  $\left\{(\mathbf{x}_i,y_i)\right\}_{i=1}^N$ with $\mathbf{x}_{i} \in \mathbb{R}^{d}$ as a natural example and $y_{i} \in\{1, \ldots, K\}$ as its corresponding label. Let $\mathcal{T}_{\boldsymbol{\theta}}$ represent a target classification model with model parameters $\boldsymbol{\theta}$. The predicted class of an input example $\mathbf{x}$ is formulated as:
\begin{equation}
\label{eq1}
\mathcal{T}_{\boldsymbol{\theta}}(\mathbf{x}_{i})=\underset{k}{\arg\max} \, T_{k}(\mathbf{x}_{i}, \boldsymbol{\theta}), \quad T_{k}(\mathbf{x}_{i}, \boldsymbol{\theta})=\exp (t_{k}(\mathbf{x}_{i}, \boldsymbol{\theta})) / \sum_{k^{*}=1}^{K} \exp (t_{k^{*}}(\mathbf{x}_{i}, \boldsymbol{\theta})),
\end{equation}
where $t_{k}(\mathbf{x}_{i}, \boldsymbol{\theta})$ is the logits output of the target model \textit{w.r.t.} class $k$, and $T_{k}(\mathbf{x}_{i}, \boldsymbol{\theta})$ approximates the probability (softmax on logits) that $\mathbf{x}_{i}$ belongs to the $k$-th class, i.e., $P(Y=k | X={x}_{i})$. We denote by $\mathbf{\tilde{x}}_{i} \in \mathbb{R}^{d}$ an adversarial example, and by $\mathbb{B}_{\epsilon}(\mathbf{x}_{i})={\mathbf\{\tilde{x}}:\left\|\mathbf{\tilde{x}}-\mathbf{x}_{i}\right\|_{p} \leq \epsilon\}$ the $L_p$ ball centered at $\mathbf{x}_{i}$ with radius $\epsilon$. 

Let $\mathcal{P}_{\boldsymbol{\phi}}$ represent a pre-processing model with model parameter $\boldsymbol{\phi}$. Given an adversarial example $\mathbf{\tilde{x}}_{i}$, the recovered example is denoted by $\mathcal{P}_{\boldsymbol{\phi}}(\mathbf{\tilde{x}}_{i})=p(\mathbf{\tilde{x}}_{i},\boldsymbol{\phi})$, where $p(\mathbf{\tilde{x}}_{i},\boldsymbol{\phi})$ is the output of the pre-processing model. We use $\mathcal{F}_{\boldsymbol{\psi}}$ to represent a full model with model parameter $\boldsymbol{\psi}$. The full network $\mathcal{F}$ is composed of a pre-processing network $\mathcal{P}$ and a target network $\mathcal{T}$. The class of an input adversarial example $\mathbf{\tilde{x}}_{i}$ predicted by $\mathcal{F}_{\boldsymbol{\psi}}$ can be represented by:
\begin{equation}
\label{eq2}
\mathcal{F}_{\boldsymbol{\psi}}(\mathbf{\tilde{x}}_{i})=\underset{k}{\arg\max} \, F_{k}(\mathbf{\tilde{x}}_{i},\boldsymbol{\psi}), \quad F_{k}(\mathbf{\tilde{x}}_{i},\boldsymbol{\psi})=\exp (f_{k}(\mathbf{\tilde{x}}_{i},\boldsymbol{\psi})) / \sum_{k^{*}=1}^{K} \exp (f_{k^{*}}(\mathbf{\tilde{x}}_{i},\boldsymbol{\psi})),
\end{equation}
where $f_{k}(\mathbf{\tilde{x}}_{i},\boldsymbol{\psi})$ is the logits output of the full model \textit{w.r.t.} class $k$, and $F_{k}(\mathbf{\tilde{x}}_{i}, \boldsymbol{\psi})$ approximates the probability that $\mathbf{\tilde{x}}_{i}$ belongs to the $k$-th class. $f_{k}(\mathbf{\tilde{x}}_{i},\boldsymbol{\psi})$ can be expressed as the output of a combined defended model: $f_{k}(\mathbf{\tilde{x}}_{i},\boldsymbol{\psi})=t_{k}(p(\mathbf{\tilde{x}}_{i},\boldsymbol{\psi}_{\phi}), \boldsymbol{\psi}_{\theta})$, where $\boldsymbol{\psi}_{\phi}$ and $\boldsymbol{\psi}_{\theta}$ respectively denote the parameters for the pre-processing network and the classification network.

\begin{figure}[t]
    \centering
    \subfigure[]{
        \label{fig2_1}
        \includegraphics[width=2.6in]{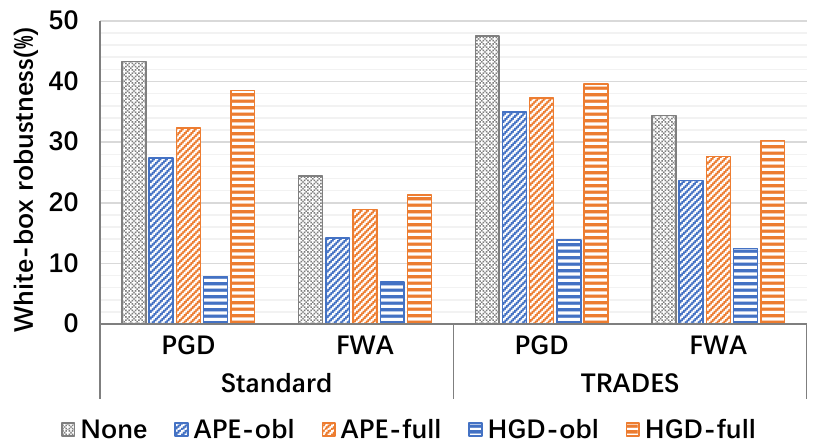}}
    \hspace{0.1in}
    \subfigure[]{
        \label{fig2_2}
        \includegraphics[width=2.6in]{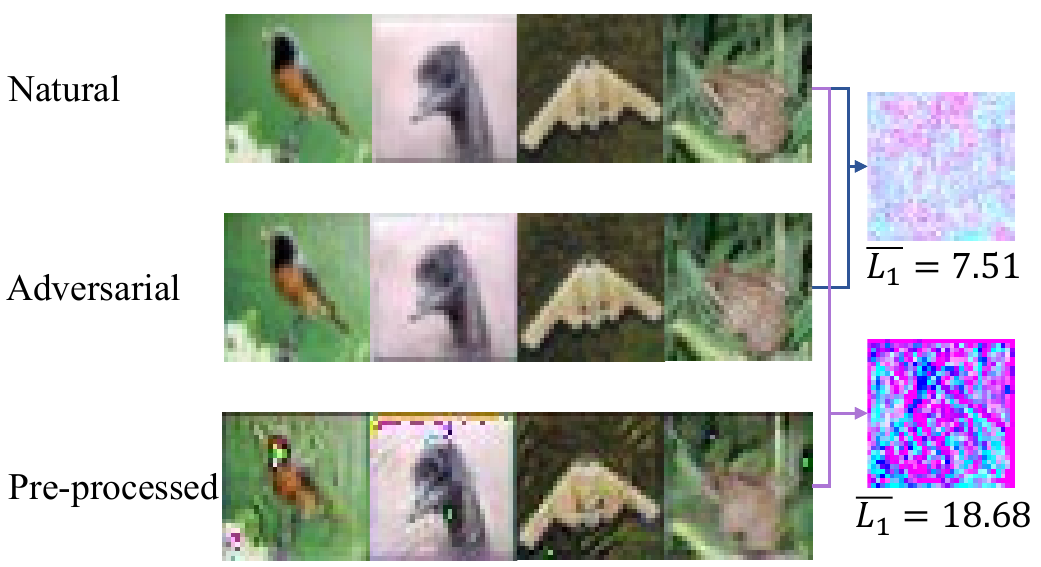}}
    \caption{(a). The distinctive influence of oblivious and full adversarial examples on \textit{CIFAR-10}. (b). A visual illustration of natural examples, adversarial examples and pre-processed examples. The adversarial examples are crafted by an adaptive PGD attack.}
\vskip -0.1in
\end{figure}

\subsection{Absence of full adversarial examples}
\label{section2.2}
Pre-processing defenses typically first exploit one or several attacks to craft adversarial examples as adversarial training examples. These training attacks are usually only applied to the target model $\mathcal{T}_{\boldsymbol{\theta}}$ without considering the pre-processing model $\mathcal{P}_{\boldsymbol{\phi}}$, and they thus are called as oblivious attacks \citep{athalye2018robustness,tramer2020adaptive}. Correspondingly, we call their adversarial examples as oblivious adversarial examples. An oblivious adversarial example $\mathbf{\tilde{x}}_{i}^{\prime}$ is crafted by solving the optimization problem:
$
\mathbf{\tilde{x}}_{i}^{\prime}= {\arg \max }_{\mathbf{\tilde{x}}_{i} \in \mathbb{B}_{\epsilon}(\mathbf{x}_{i})} \mathbbm{1}(\mathcal{T}_{\boldsymbol{\theta}}(\mathbf{\tilde{x}}_{i}) \neq y_{i}), 
$
where $\mathbbm{1}(\cdot)$ denotes the indicator function. 

However, adaptive attacks tend to consider that a pre-processing defense model has been deployed. They can optimize a combined loss function on $\mathcal{T}_{\boldsymbol{\theta}}(\mathcal{P}_{\boldsymbol{\phi}}(\mathbf{x}_i))$ \citep{tramer2020adaptive} to craft adaptive adversarial examples. Unfortunately, for most pre-processing defenses, their oblivious adversarial examples are specific to the fixed $\mathcal{T}_{\boldsymbol{\theta}}$ and not associated with $\mathcal{P}_{\boldsymbol{\phi}}$. The influence of the gradient passing through $\mathcal{P}_{\boldsymbol{\phi}}$ on the adversarial robustness of $\mathcal{P}_{\boldsymbol{\phi}}$ is ignored. From this perspective, using such training examples cannot guarantee that the defense model can effectively deal with the adaptive attacks. To solve this problem, a natural idea is to use full adversarial examples which are crafted against the full model to replace oblivious adversarial examples. To prove this idea, we investigate the influence of full adversarial examples on the robustness and find that they indeed exhibit a better impact compared to oblivious adversarial examples. Specifically, we conduct a proof-concept experiment on \textit{CIFAR-10} to present the distinctive influence of oblivious and full adversarial examples.

We consider two pre-processing defense methods, APE-G and HGD, which have great performances against oblivious attacks. The pre-processing models are trained by using the $L_{\infty}$ PGD with the original training strategies in \citep{shen2017ape,liao2018defense}. Given two robust target models which are adversarially trained via \textit{Standard} \citep{madry2017towards} and \textit{TRADES} \citep{zhang2019theoretically}, we evaluate the white-box robustness of the original pre-processing models (-obl) by using white-box adaptive attacks (see the blue bars in Fig~\ref{fig2_1}). The details of these test attacks can be found in Section~\ref{section4.2}. We then use same $L_{\infty}$ PGD to craft full adversarial examples for retraining the pre-processing models.: 
\begin{equation}
\label{eq3}
\mathbf{\tilde{x}}_{i}^{*}={\arg \max }_{\mathbf{\tilde{x}}_{i}} \mathbbm{1}(\mathcal{T}_{\boldsymbol{\theta}}(\mathcal{P}_{\boldsymbol{\phi}^{\prime}}(\mathbf{\tilde{x}}_{i})) \neq y_{i}), \, \mathbf{\tilde{x}}_{i} \in \mathbb{B}_{\epsilon}(\mathbf{x}_{i}),
\end{equation} 
where $\phi^{\prime}$ denotes the parameters of the retrained pre-processing models. We use the same loss functions as those in their original papers, and conduct an adversarial training with 40 iterations. The white-box robustness of the retrained pre-processing model is significantly better than that of original models (see the orange bars in Fig~\ref{fig2_1}). The results of this proof-concept experiment demonstrate that using full adversarial examples could improve the white-box robustness of the pre-processing defense. This discovery inspired us to use full adversarial examples to train the pre-processing model.

\subsection{Vulnerability of the pre-processing model}
\label{section2.3}
Note that the results in Figure~\ref{fig2_1} also show that simply modifying the adversarial training examples in pre-processing methods does not completely alleviate the robustness degradation effect. This may be caused by the vulnerability of the pre-processing model. An attacker can mislead the pre-processing model to generate an output with respect to wrong class by disrupting the pre-processed example as much as possible. As shown in Fig~\ref{fig2_2}, the $L_{1}$ distance between the adversarial examples and the natural examples is 7.51 on average, whereas the distance between the pre-processed examples and natural examples is 18.68 on average. The pre-processed examples have more undesirable noise than the adversarial examples, which indicated that the inherent robustness of the pre-processing model is not sufficient. Therefore, the approach for training a pre-processing defense model should involve the adversarial risk of the pre-processing model to enhance its white-box robustness.

\begin{figure}[t]
    \centering
    \label{fig3}
    \includegraphics[width=5in]{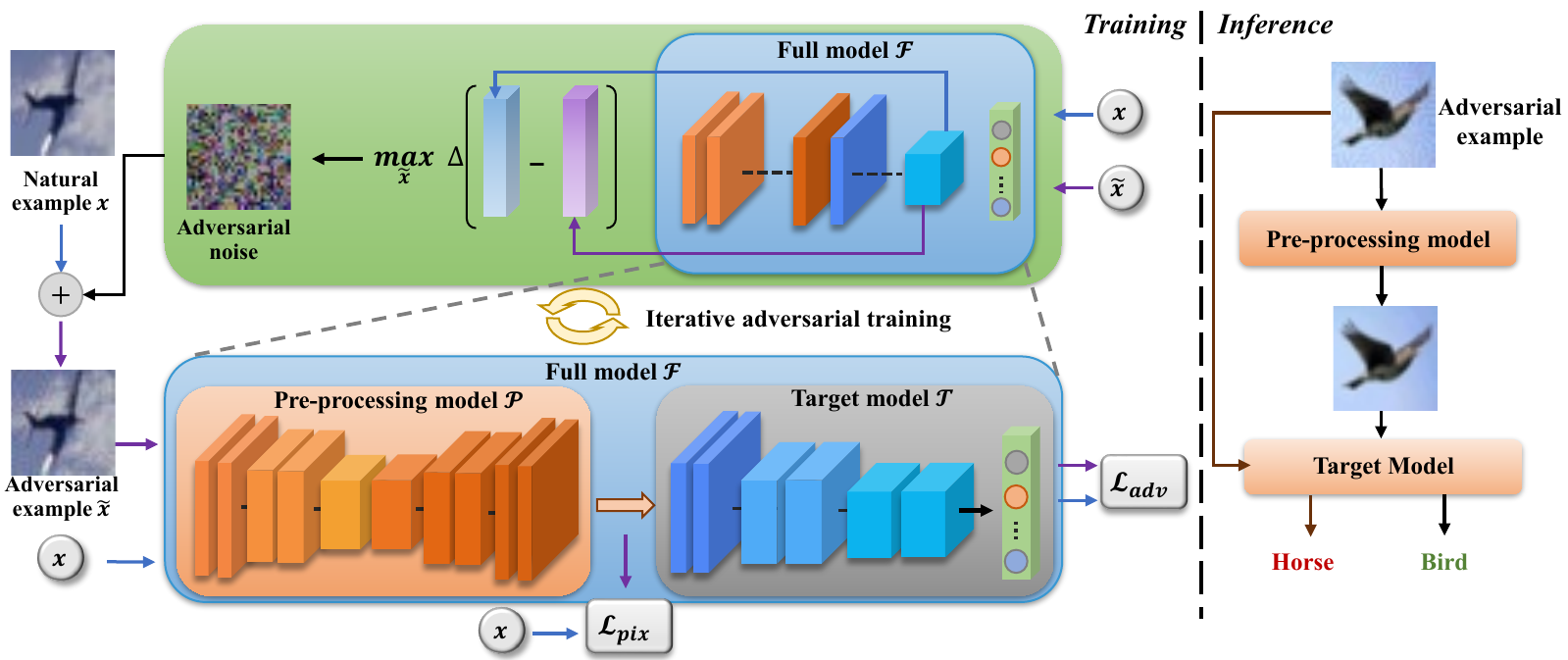}
    \caption{A visual illustration of our \textit{Joint Adversarial Training based Pre-processing} (JATP) defense. We use adversarial examples against the full model $\mathcal{F}$ to train a pre-processing model $\mathcal{P}$ that minimizes a hybrid loss composed of the pixel-wise loss $\mathcal{L}_{\mathbf{pix}}$ and the adversarial loss $\mathcal{L}_{\mathbf{adv}}$.}
\vskip -0.1in
\end{figure}

Since a white-box adaptive attack can exploit the knowledge of the full model to break the defense, 
we exploit the full adversarial examples to formulate an adversarial risk for the pre-processing model. In addition, \citet{wang2019improving} propose to explicitly differentiate the misclassified and correctly classified examples during the training, and they design a misclassification aware adversarial risk for improving the adversarial robustness of a target model. Inspired by their works, we formulate the adversarial risk of the pre-processing model as:
\begin{equation}
\label{eq4}
\mathcal{R}(\mathcal{P}_{\boldsymbol{\psi}_{\phi}}) = \frac{1}{N} \sum_{i=1}^{N}\{\mathbbm{1}(\mathcal{F}_{\boldsymbol{\psi}}(\mathbf{\tilde{x}}_{i}^{*}) \neq y_{i})+\mathbbm{1}(\mathcal{F}_{\boldsymbol{\psi}}(\mathbf{x}_{i}) \neq \mathcal{F}_{\boldsymbol{\psi}}(\mathbf{\tilde{x}}_{i}^{*})) \cdot \mathbbm{1}(\mathcal{F}_{\boldsymbol{\psi}}(\mathbf{x}_{i}) \neq y_{i})\},
\end{equation}
where $\mathcal{F}_{\boldsymbol{\psi}}(\mathbf{x})=\mathcal{T}_{\boldsymbol{\psi}_{\theta}}(\mathcal{P}_{\boldsymbol{\psi}_{\phi}}(\mathbf{x}))$ is the class of the input $\mathbf{x}$ predicted by the full model $\mathcal{F}_{\boldsymbol{\psi}}$, and $\mathbf{\tilde{x}}_{i}^{*}={\arg \max }_{\mathbf{\tilde{x}}_{i}\in \mathbb{B}_{\epsilon}(\mathbf{x}_{i}) } \mathbbm{1}(\mathcal{F}_{\boldsymbol{\psi}}(\mathbf{\tilde{x}}_{i}) \neq y_{i})$ denotes the full adversarial example. The $\mathbbm{1}(\mathcal{F}_{\boldsymbol{\psi}}(\mathbf{x}_{i}) \neq \mathcal{F}_{\boldsymbol{\psi}}(\mathbf{\tilde{x}}_{i}^{*})) \cdot \mathbbm{1}(\mathcal{F}_{\boldsymbol{\psi}}(\mathbf{x}_{i}) \neq y_{i})$ is the misclassification aware regularization term. Note that the target model $\mathcal{T}_{\boldsymbol{\psi}_{\theta}}$ is pre-trained and its model parameters $\boldsymbol{\psi}_{\theta}$ is fixed during the training process. According to this adversarial risk, we could design a method to train a pre-processing defense model with better inherent adversarial robustness.

\section{Proposed method}
\label{section3}
Motivated by the analyses in the previous section, we propose a \textit{Joint Adversarial Training based Pre-processing} (JATP) defense to mitigate the robustness degradation effect. We use full adversarial examples to train a pre-processing model that minimizes a hybrid loss composed of a pixel-wise loss and an adversarial loss. Fig~\ref{fig3} shows the visual illustration of our proposed defense. 

\subsection{Adversarial training examples}
\label{section3.1}
Based on the proof-concept experiment in Section~\ref{section2.2}, we apply adversarial attacks to the full model $\mathcal{F}_{\boldsymbol{\psi}}$ to craft full adversarial examples for training the pre-processing model $\mathcal{P_{\boldsymbol{\psi}_{\phi}}}$. In addition, \citet{kurakin2016adversarial} point out that directly maximizing the cross-entropy loss in label space to craft adversarial training examples may lead to the \textit{label leakage} problem. The problem allows a defense model to overfit on specific perturbations thus affecting model generalization to unseen attacks. \citet{zhang2019defense} show that using a highly transferable attack to craft adversarial examples is beneficial to improve the generalization of the defense model, and finding the worst-case distortion in feature space can strength the transferability of an adversarial attack. Therefore, we redesign the adversarial training examples as follows:
\begin{equation}
\label{eq5}
\mathbf{\tilde{x}}_{i}^{*}= {\arg \max }_{\mathbf{\tilde{x}}_{i}} d(f^{*}(\mathbf{\tilde{x}}_{i}, {\boldsymbol{\psi}}),f^{*}(\mathbf{x}_{i},{\boldsymbol{\psi}})), \, \mathbf{\tilde{x}}_{i} \in \mathbb{B}_{\epsilon}(\mathbf{x}_{i}),
\end{equation}
where $f^{*}(\mathbf{x},{\boldsymbol{\psi}})$ denotes the feature map of $\mathbf{x}$ on an internal layer of $\mathcal{F}_{\boldsymbol{\psi}}$, and $d(\cdot)$ is a distance metric such as $L_2$, cosine similarity or Wasserstein distance \citep{arjovsky2017wasserstein}. In this paper, we use the ResNet-18 \citep{he2016deep} as the network of target model during training. Since the bottom layers tend to learn low-level features while the deeper ones are too specific to the label space, we choose the last convolution layer in the third basic block to obtain the feature map. We also study the influences of feature maps on different internal layers on the white-box robustness, and present their results in supplementary material B. 

\subsection{Loss function for JATP defense}
\label{section3.2}
We exploit a hybrid loss function to train our proposed JATP defense. The loss function is composed of two terms: pixel-wise loss and adversarial loss. 

\textbf{Pixel-wise loss.}
Our defense method aims to remove adversarial noise and preserve the original information of natural examples. Thus, we expect the recovered examples to be as close to the natural examples as possible. In addition, due to the similarity between the adversarial pattern and noise, smoothing examples can be helpful in reducing the interference of adversarial patterns. We apply a $L_2$ loss to measure the distance between recovered examples and natural examples in pixel space:
\begin{equation}
\label{eq6}
\mathcal{L}_{\mathbf{pix}}(\boldsymbol{\psi}_{\phi})=\left\|\mathcal{P}_{\boldsymbol{\psi}_{\phi}}(\mathbf{\tilde{x}}^{*})-\mathbf{x}\right\|_{2}.
\end{equation}

We study the influence of this pixel-wise loss on our proposed defense in Section~\ref{section4.3}, this loss help our pre-processing model retain the color and texture of natural examples, which can enhance the transferability of our defense across different target models.

\textbf{Adversarial loss.}
As presented in Section~\ref{section2.3}, the adversarial risk of the pre-processing model consists of two parts: (1) the standard adversarial term $\mathbbm{1}(\mathcal{F}_{\boldsymbol{\psi}}(\mathbf{\tilde{x}}_{i}^{*}) \neq y_{i})$ and (2) the regularization adversarial term $\mathbbm{1}(\mathcal{F}_{\boldsymbol{\psi}}(\mathbf{x}_{i}) \neq \mathcal{F}_{\boldsymbol{\psi}}(\mathbf{\tilde{x}}_{i}^{*})) \cdot \mathbbm{1}(\mathcal{F}_{\boldsymbol{\psi}}(\mathbf{x}_{i}) \neq y_{i})$, both of which have the indicator function. In practice, it is intractable to directly optimize the 0-1 loss of the indicator function, and we thus use appropriate loss functions to replace the 0-1 loss.

For the standard adversarial term $\mathbbm{1}(\mathcal{F}_{\boldsymbol{\psi}}(\mathbf{\tilde{x}}_{i}^{*}) \neq y_{i})$, inspired by \citep{wang2019improving}, we use a boosted cross-entropy (BCE) loss to replace the 0-1 loss, instead of the commonly used CE loss \citep{madry2017towards,liao2018defense}. The BCE loss is defined as:
\begin{equation}
\label{eq7}
\ell_{\operatorname{BCE}}(\mathbf{\tilde{x}}_{i}^{*},\boldsymbol{\psi}_{\phi},y_i)= -\log t_{y_i}(p(\mathbf{\tilde{x}}_{i}^{*},\boldsymbol{\psi}_{\phi}),\boldsymbol{\psi}_{\theta}) -\log (1-\max _{k \neq y_{i}} t_{k}(p(\mathbf{\tilde{x}}_{i}^{*},\boldsymbol{\psi}_{\phi}),\boldsymbol{\psi}_{\theta})),
\end{equation}
where $t_{k}(\cdot)$ is the probability output defined in Eq~\ref{eq1}, and $p(\cdot)$ is the pre-processed output defined in Eq~\ref{eq2}. $\boldsymbol{\psi}_{\theta}$ is the fixed parameters of the pre-trained target model $\mathcal{T}_{\boldsymbol{\psi}_{\theta}}$. The first term is the commonly used cross-entropy loss, and the second term is used to reduce undesirable distortions generated by $\mathcal{P}_{\boldsymbol{\psi}_{\phi}}$ for decreasing the probability of the pre-processed examples with respect to the wrong classes.

For the regularization term, $\mathcal{F}_{\boldsymbol{\psi}}(\mathbf{x}_{i}) \neq \mathcal{F}_{\boldsymbol{\psi}}(\mathbf{\tilde{x}}_{i}^{*})$ implies that adversarial examples have different output distribution to that of natural examples. Considering that the output is directly manipulated by the feature map on the internal layers of $\mathcal{F}_{\boldsymbol{\psi}}$, and the full adversarial examples used to train the pre-processing model are found in the feature space, we thus use a \textit{feature similarity metric} (FSM) to replace the first indicator function $\mathbbm{1}(\mathcal{F}_{\boldsymbol{\psi}}(\mathbf{x}_{i}) \neq \mathcal{F}_{\boldsymbol{\psi}}(\mathbf{\tilde{x}}_{i}^{*}))$. The first term is expressed as: 
\begin{equation}
\label{eq8}
\ell_{\operatorname{FSM}}(\mathbf{\tilde{x}}_{i}^{*},\mathbf{x},\boldsymbol{\psi}_{\phi})= d(t^{*}(p(\mathbf{\tilde{x}}_{i},\boldsymbol{\psi}_{\phi}),\boldsymbol{\psi}_{\theta}),t^{*}(p(\mathbf{x}_{i}^{*},\boldsymbol{\psi}_{\phi}),\boldsymbol{\psi}_{\theta})),
\end{equation}
where $d(\cdot)$ is a mean square error metric, and $t^{*}(\cdot)$ is the feature map on the internal layer of $\mathcal{T}_{\boldsymbol{\psi}_{\theta}}$ defined in Section~\ref{section3.1}. The another indicator function $\mathbbm{1}(\mathcal{F}_{\boldsymbol{\psi}}(\mathbf{x}_{i}) \neq y_{i})$ is a condition that emphasizes learning on misclassified examples \citep{wang2019improving}. The misclassification based constrain used in adversarial training also help to enhance the inherent robustness of the pre-processing model. We use a soft decision scheme, \textit{i.e.,} the output probability $1-t_{y_i}(p(\mathbf{x}_{i},{\boldsymbol{\psi}_{\psi}}),{\boldsymbol{\psi}_{\theta}})$ to replace the indicator function $\mathbbm{1}(\mathcal{F}_{\boldsymbol{\psi}}(\mathbf{x}_{i}) \neq y_{i})$. The overall adversarial loss of the pre-processing model can be defined as follows:
\begin{equation}
\label{eq9}
\mathcal{L}_{\mathbf{adv}}(\boldsymbol{\psi}_{\phi})= \frac{1}{N}\sum_{i=1}^{N} \{\ell_{\operatorname{BCE}}(\mathbf{\tilde{x}}_{i}^{*},\boldsymbol{\psi}_{\phi},y_i) + \alpha \cdot \ell_{\operatorname{FSM}}(\mathbf{\tilde{x}}_{i}^{*},\mathbf{x},\boldsymbol{\psi}_{\phi}) \cdot (1-t_{y_i}(p(\mathbf{x}_{i},{\boldsymbol{\psi}_{\psi}}),{\boldsymbol{\psi}_{\theta}}))\}, 
\end{equation}
where $\alpha$ is a positive hyperparameter.

\textbf{The overall loss.}
Based on the pixel-wise loss and the adversarial loss, we present the overall loss for our proposed JATP defense:
\begin{equation}
\label{eq10}
\mathcal{L}_{\mathbf{JATP}}(\boldsymbol{\psi}_{\phi})= \mathcal{L}_{\mathbf{adv}}(\boldsymbol{\psi}_{\phi}) + \beta \cdot \mathcal{L}_{\mathbf{pix}}(\boldsymbol{\psi}_{\phi}),
\end{equation}
where $\beta$ is a hyperparameter. The overall procedure is summarized in Algorithm~\ref{alg1}.

\begin{algorithm}[t]
   \caption{JATP: Joint adversarial training based Pre-processing defense}
   \label{alg1}
\begin{algorithmic}[1]
   \REQUIRE A pre-trained target model $\mathcal{T}_{\boldsymbol{\psi}_{\theta}}$, a pre-processing model $\mathcal{P}_{\boldsymbol{\psi}_{\phi}}$, batch of natural examples $\mathbf{x}$, perturbation budget $\epsilon$, number of iterations $T$.
   \STATE Initialization;
   \FOR{$i=1$ to $T$}
   \STATE Craft full adversarial example $\tilde{\mathbf{x}}^{*}$ at the given perturbation budget $\epsilon$ using Eq.~\ref{eq5};
   \STATE Forward-pass $\tilde{\mathbf{x}}^{*}$ through $\mathcal{P}_{\boldsymbol{\psi}_{\phi}}$ and calculate $\mathcal{L}_{\mathbf{pix}}$ using Eq.~\ref{eq6};
   \STATE Forward-pass $\mathcal{P}_{\boldsymbol{\psi}_{\phi}}(\tilde{\mathbf{x}}^{*})$ through $\mathcal{T}_{\boldsymbol{\psi}_{\theta}}$ and calculate $\mathcal{L}_{\mathbf{adv}}$ using Eq.~\ref{eq9};
   \STATE Back-pass and update $\mathcal{P}_{\boldsymbol{\psi}_{\phi}}$ to minimize $\mathcal{L}_{\mathbf{JATP}}$ using Eq.~\ref{eq10};
   \ENDFOR
   \RETURN $\boldsymbol{\psi}_{\theta}$.
\end{algorithmic}
\end{algorithm}

\section{Experiments}
\label{section4}
In this section, we first introduce the experimental setup used in this paper (Section~\ref{section4.1}). Then, we evaluate the adversarial robustness of our proposed JATP defense against unseen types of adaptive attacks (Section~\ref{section4.2}). Finally, we conduct the ablation and sensitivity studies to provide a further understanding of our defense method (Section~\ref{section4.3}).

\subsection{Experiment setup}
\label{section4.1}
\textbf{Datasets.} We evaluate the adversarial robustness of pre-processing defenses on two popular benchmark datasets, i.e., \textit{SVHN} \citep{netzer2011reading} and \textit{CIFAR-10} \citep{krizhevsky2009learning}. \textit{SVHN} and \textit{CIFAR-10} both have 10 classes of images, but the former contains 73,257 training images and 26,032 test images, and the latter contains 60,000 training images and 10,000 test images. All natural images are normalized into [0,1], and are performed simple data augmentations including 4-pixel padding with $32 \times 32$ random crop and random horizontal flip. 

\textbf{Target model settings.} For both \textit{SVHN} and \textit{CIFAR-10}, target models are built on a ResNet-18 architecture \cite{he2016deep}. Four adversarial training strategies are utilized to pre-train the target models, including \textit{Standard} \citep{madry2017towards} \textit{MMA} \citep{ding2019sensitivity}, \textit{TRADES} \citep{zhang2019theoretically} and \textit{MART} \citep{wang2019improving}. Hyperparameters of these strategies are configured as per their original papers: $\lambda= 6$ for \textit{TRADES} and \textit{MART}. A PGD-10 attack with random start and step size 0.007 is utilized as the training attack. The perturbation budget is $\epsilon=8/255$ for both \textit{SVHN} and \textit{CIFAR-10}. The link to the implementation code of target models can be found in supplementary material C.

\textbf{Defense Settings.} We use three pre-processing defense methods as the baselines: APE-G \citep{shen2017ape}, HGD \citep{liao2018defense} and NRP \citep{naseer2020self}. The architectures of the pre-processing model in APE-G and HGD are same as those in their original papers. We use the classification model adversarially trained by \textit{TRADES} as the the target model during the training process. A PGD-40 attack with step size 0.007 is used as the training attack for APE-G and HGD. The perturbation budget $\epsilon$ is set to $8/255$ for both \textit{SVHN} and \textit{CIFAR-10}. For NRP, we use the basic block in the original paper and only reduce the number of basic blocks to 3. The training strategies and other hyperparameters of baselines are consistent with the settings on \textit{CIFAR-10} in their original papers. For our method, we utilize the same architecture as NPR to build our pre-processing network. Our pre-processing model is trained using SGD \citep{andrew2007scalable} with momentum 0.9, weight decay $2\times10^{-4}$ and an initial learning rate of $10^{-2}$, which is divided by 10 at the $15-th$ and $25-th$ epoch (30 epochs in total). $\epsilon$ is set to $8/255$ and the step size is 0.01. The hyperparameters $\alpha$ and $\beta$ are set to 5.0 and 3.0 respectively. For fair comparison, all experiments are conduced on four NVIDIA RTX 2080 GPUs, and all methods are implemented by PyTorch.

\subsection{Adversarial robustness evaluation}
\label{section4.2}

\begin{table*}[t]
\caption{White-box robustness (percentage) of pre-processing defenses on \textit{CIFAR-10}. We show the most successful pre-processing defense with \textbf{bold}. The target model is trained by \textit{TRADES}.}
\label{tab1}
\renewcommand\tabcolsep{3pt}
\begin{center}
\begin{tabular}{c|cccccc}
\hline
Attack & Natural & PGD & DLR & AA & FWA & TI-BIM \\ \hline
None & 78.45 $\pm$ 0.06 & 47.48 $\pm$ 0.05 & 45.02 $\pm$ 0.01 & 44.10 $\pm$ 0.07 & 34.39 $\pm$ 0.06 & 59.67 $\pm$ 0.02 \\
APE-G & 78.40 $\pm$ 0.08 & 34.94 $\pm$ 0.07 & 33.11 $\pm$ 0.01 & 16.35 $\pm$ 0.06 & 23.71 $\pm$ 0.03 & 57.80 $\pm$ 0.04 \\
HGD & 78.17 $\pm$ 0.08 & 13.84 $\pm$ 0.03 & 12.66 $\pm$ 0.01 & 17.36 $\pm$ 0.07 & 12.5 $\pm$ 0.01 & 56.90 $\pm$ 0.05 \\
NRP & 78.38 $\pm$ 0.08 & 24.28 $\pm$ 0.08 & 22.94 $\pm$ 0.07 & 18.51 $\pm$ 0.05 & 11.78 $\pm$ 0.08 & 57.40 $\pm$ 0.07 \\
JATP & \textbf{79.00 $\pm$ 0.03} & \textbf{46.14 $\pm$ 0.01} & \textbf{43.54} $\pm$ 0.03 & \textbf{43.25 $\pm$ 0.04} & \textbf{31.57 $\pm$ 0.05} & \textbf{60.04 $\pm$ 0.06} \\ \hline
\end{tabular}
\end{center}
\vskip -0.15in
\end{table*}

\begin{table*}[t]
\caption{White-box robustness (percentage) of pre-processing defenses on different target models. We show the most successful pre-processing defense with \textbf{bold}  and the second one with \uline{underline}.}
\label{tab2}
\renewcommand\tabcolsep{2.8pt}
\begin{center}
\begin{tabular}{c|cccc|cccc|cccc}
\hline
Target & \multicolumn{4}{c|}{Standard} & \multicolumn{4}{c|}{MMA} & \multicolumn{4}{c}{MART} \\ 
Attack & Natural & PGD & DLR & FWA & Natural & PGD & DLR & FWA & Natural & PGD & DLR & FWA \\ \hline
None & 82.38 & 43.29 & 43.27 & 24.39 & 82.88 & 43.56 & 42.60 & 21.97 & 78.19 & 48.54 & 44.90 & 34.28 \\
APE-G & 82.54 & 27.35 & 30.68 & 14.17 & \uline{82.53} & 29.02 & 30.59 & 13.18 & 78.28 & 34.34 & 32.15 & 21.28 \\
HGD & \uline{82.64} & 7.86 & 10.23 & 6.97 & 82.17 & 9.09 & 10.29 & 7.24 & 77.74 & 10.97 & 11.31& 10.09 \\
NRP & 82.43 & 13.71 & 17.72 & 5.01 &82.52 & 16.56 & 18.46& 4.50 & 77.95 & 21.32 & 20.26 & 9.64 \\
JATP & \textbf{83.65} & \uline{41.65} & \uline{42.75} & \uline{22.07} & \textbf{83.14} & \uline{41.73} & \uline{40.89} & \uline{19.29} & \textbf{79.25} & \uline{46.75} & \uline{43.29} & \uline{31.54} \\
JATP$^{\prime}$ & 82.45& \textbf{43.89} & \textbf{42.99} & \textbf{24.86} & 82.29 & \textbf{42.82} & \textbf{41.64} & \textbf{19.82} & \uline{77.89} & \textbf{48.29} & \textbf{45.00} & \textbf{35.04} \\\hline
\end{tabular}
\end{center}
\vskip -0.1in
\end{table*}

\begin{figure}[t]
    \centering
    \subfigure[BPDA]{
        \label{fig4_1}
        \includegraphics[width=2.6in]{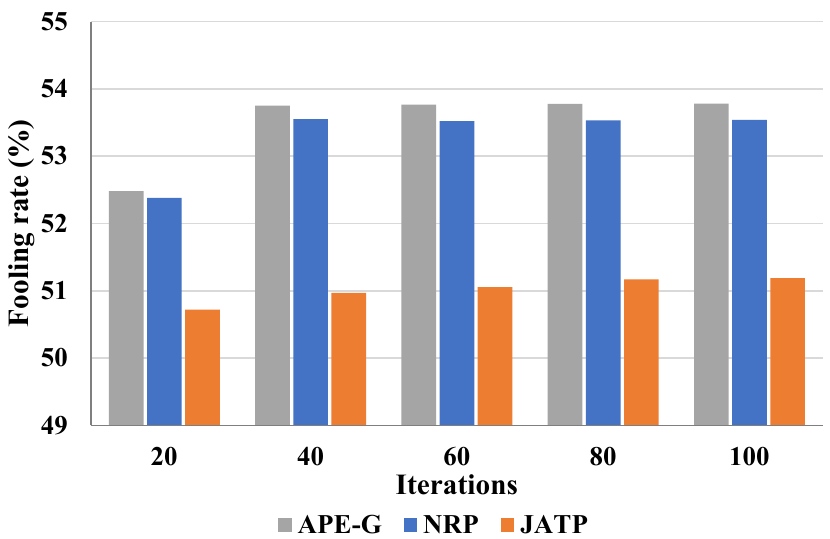}}
    \hspace{0.1in}
    \subfigure[Ablation study]{
        \label{fig4_2}
        \includegraphics[width=2.6in]{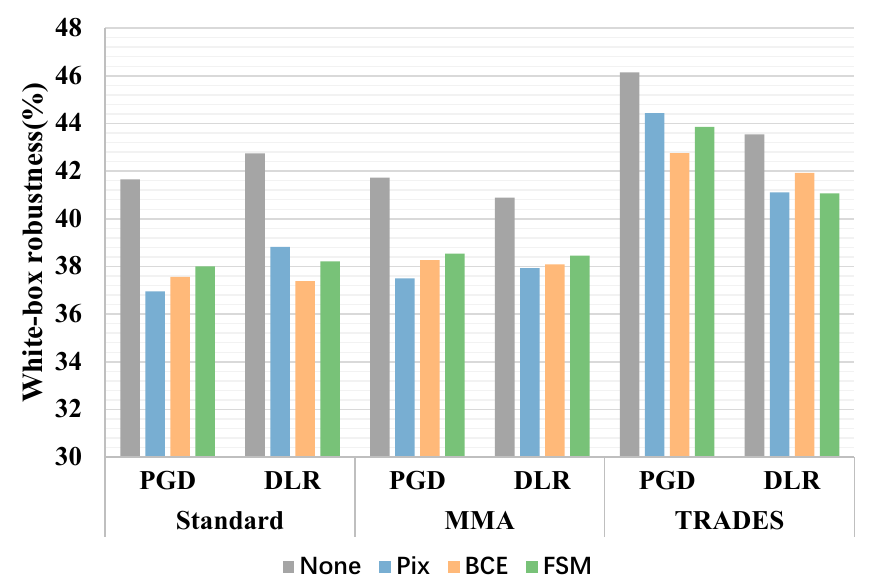}}
    \caption{(a). Fooling rate (\textit{lower is better}) of BPDA and PGD against pre-processing models. The target model is trained by \textit{TRADES}. (b). Ablation study. We remove the pixel-wise loss ("Pix"), BCE adversarial loss ("BCE") and feature similarity adversarial loss ("FSM") respectively to investigate their impacts on our model.}
\vskip -0.15in
\end{figure}

\textbf{Robustness against adaptive attacks.}
We first evaluate the robustness of all pre-processing defense models against five types of attacks for both \textit{SVHN} and \textit{CIFAR-10}: PGD (40-step $L_{\infty}$ PGD with step size 0.01), DLR (40-step $L_{\infty}$ PGDDLR \citep{croce2020reliable} with step size 0.007), AA ($L_{\infty}$-norm version of AutoAttack), FWA (20-step FWA with step size 0.01) and TI-BIM \citep{dong2019evading,xie2019improving}. The perturbation budget is set to 8/255. All attacks exploit an adaptive attack strategy that the attacker have full access to the architectures and model parameters of both the pre-processing model and the target model. The white-box robustness of all defense models are shown in Table~\ref{tab1}. Our proposed pre-processing defense JATP achieves the best robustness against all five types of attacks, and significantly mitigate the robustness degradation effect compared with other pre-processing models. Due to the space limitation, we present the results on \textit{SVHN} in the supplementary material C. In addition, we use an adaptive attack strategy BPDA \citep{athalye2018obfuscated} to detect whether our work relies heavily on \textit{obfuscated gradients}. We combine BPDA with $L_{\infty}$ PGD to bypass the pre-processing defenses. As shown in Figure~\ref{fig4_1}, our method achieves lower fooling rates than other baselines.

\textbf{Cross-model defense.}
We apply our proposed JATP defense to other three target models (\textit{i.e.,} \textit{Standard}, \textit{MMA} and \textit{MART}) to evaluate its transferability. The white-box robustness of our proposed JATP defense on \textit{CIFAR-10} is reported in Table~\ref{tab2}. We present the standard deviation in supplementary material C. Again, our proposed defense achieves higher robustness than other pre-processing defenses. Furthermore, we conduct an additional experiment under a relaxed constrain. That is, we update the parameters of the target model and the pr-processing model together during the training process, instead of using a pre-trained target model and fixing its model parameters. We then transfer the obtained pre-processing model denoted by JATP$^{\prime}$ to above three target models. We find that the transferability of JATP$^{\prime}$ defense has a slight improvement compared with JATP defense.

\subsection{Ablation study}
\label{section4.3}
In this section, we conduct an ablation study to further understand the proposed JATP defense. We respectively remove the pixel-wise loss, BCE adversarial loss and feature similarity adversarial loss to investigate their impacts on our model. A target model trained by \textit{TRADES} is used during the training process. As illustrated in Figure~\ref{fig4_2}, removing pixel-wise loss mainly affects the transferability of the pre-processing model across different target models. Removing the BCE adversarial loss and the feature similarity adversarial loss would lead to a significant robustness degradation, which shows the adversarial loss is important to the pre-processing for improving its adversarial robustness.

\section{Conclusion}
\label{section5}
In this paper, we analyze two potential causes of robustness degradation effect: (1) adversarial training examples used in pre-processing methods are independent to the pre-processing model; and (2) the adversarial risk of the pre-processing model is neglected during the training process. To solve this problem, we first formulate a feature similarity based adversarial risk for the pre-processing model to improve its inherent robustness by using full adversarial examples crafted in a feature space. We then introduce a pixel-wise loss to improve the cross-model transferability of the pre-processing model. Based on these, We propose a \textit{Joint Adversarial Training based Pre-processing} (JATP) defense to minimize the overall risk in a dynamic manner. Experimental results show that our method effectively mitigates the robustness degradation effect against in comparison to previous pre-processing defenses. We hope that the work in this paper can provide some inspiration for future pre-processing defenses in improving the white-box robustness. The limitation of our work is that the our proposed method is only suitable for input denoising defenses so far, and we have not explored how to apply it to other pre-processing defenses. In the future, we plan to combine the pre-processing defense with recently proposed robustness models \citep{wu2020adversarial,chen2021robust} and explore the potential improvements on the white-box robustness of pre-processing defenses.

\newpage

\normalem

{\small
\bibliographystyle{plainnat}
\bibliography{egbib}
}


\end{document}